# WarpAdam: A new Adam optimizer based on Meta-Learning approach

Chengxi Pan[1,*], Junshang Chen[2], Jingrui Ye[3]

[1]Beijing University of Chemistry Technology, sweetchris1196@gmail.com
[2]Hohai University, ChenJunShang@hotmail.com
[3]FuZhou University, JINGRUI.YE.2022@MUMAIL.IE
*Corresponding author email: sweetchris1196@gmail.com

**Abstract:** Optimal selection of optimization algorithms is crucial for training deep learning models. The Adam optimizer has gained significant attention due to its efficiency and wide applicability. However, to enhance the adaptability of optimizers across diverse datasets, we propose an innovative optimization strategy by integrating the "warped gradient descend" concept from Meta Learning into the Adam optimizer. In the conventional Adam optimizer, gradients are utilized to compute estimates of gradient mean and variance, subsequently updating model parameters. Our approach introduces a learnable distortion matrix, denoted as $P$, which is employed for linearly transforming gradients. This transformation slightly adjusts gradients during each iteration, enabling the optimizer to better adapt to distinct dataset characteristics. By learning an appropriate distortion matrix $P$, our method aims to adaptively adjust gradient information across different data distributions, thereby enhancing optimization performance. Our research showcases the potential of this novel approach through theoretical insights and empirical evaluations. Experimental results across various tasks and datasets validate the superiority of our optimizer that integrates the "warped gradient descend" concept in terms of adaptability. Furthermore, we explore effective strategies for training the adaptation matrix $P$ and identify scenarios where this method can yield optimal results. In summary, this study introduces an innovative approach that merges the "warped gradient descend" concept from Meta Learning with the Adam optimizer. By introducing a learnable distortion matrix $P$ within the optimizer, we aim to enhance the model's generalization capability across diverse data distributions, thus opening up new possibilities in the field of deep learning optimization.

## 1 Introduction

Meta-learning, or "learning to learn," involves infer- ring effective learning strategies from past experiences to facilitate rapid adaptation to new tasks [10]. In meta- learning, the choice of optimizer significantly influences the efficiency and effectiveness of the learning process. An optimizer updates model parameters during training to minimize the loss function. In the context of meta- learning, where fast adaptation is crucial, a well-designed optimizer can greatly impact generalization and stability.

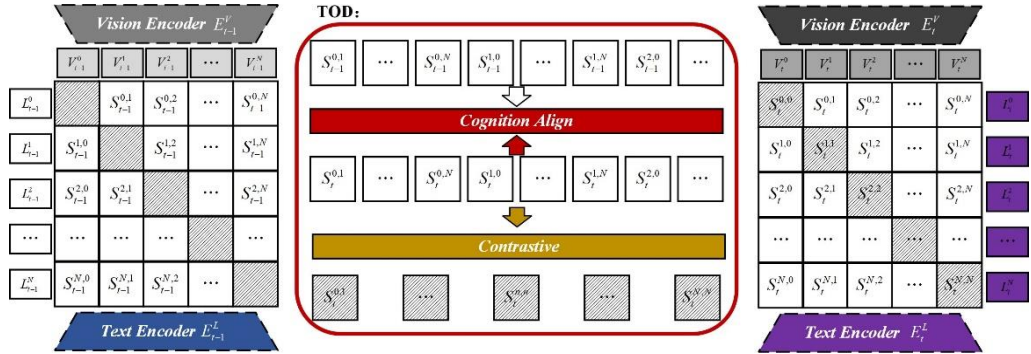

**Figure 1:** TOD - Transfer off-diagonal information- Matrix. Alleviate the cognitive confusion of the model during continuous training by limiting the off-diagonal information distribution during the current model update process.

However, conventional optimizers like SGD (Stochastic Gradient Descent) and Adam face challenges [4] [11]. SGD's sensitivity to learning rate and step size can lead to slow convergence and local minima issues, while Adam's instability and hyperparameter sensitivity hinder its effectiveness in meta-learning [32]. Moreover, Adam's memory usage for momentum and moment estimates poses challenges, particularly for large models [28].

In this context, the optimizer's role becomes crucial due to rapid adaptation requirements. It must efficiently update parameters while ensuring stability and quick con- vergence, with a bias for strong generalization [34].

To address these challenges, we introduce "Warp- Grad," a novel method that combines memory-based and gradient-based approaches to optimize meta-learning. It preconditions gradients using warped gradient concepts, rectifying Adam's exponential moving average issues and addressing convergence challenges in mini-batches with significant low-frequency gradients.

Traditional Adam combines RMSprop and Momen- tum, estimating first and second moments using exponential moving averages. Yet, it neglects potential interrelations between moments, impacting optimization efficacy. Additionally, initial iterations exhibit biases in first and second moments, requiring bias correction for smoother iteration [30] [29].

We propose a Framework with a "warped layer" introducing non-linearity to make preconditioning data- dependent. That is TOD-Transfer off-diagonal information-Matrix. This adaptiveness enhances preconditioning based on specific data characteristics. Unlike previous works constrained by block-diagonal structures, our approach preserves gradient descent's convergence properties. Moreover, meta-learning through warped lay- ers captures task distribution characteristics for better performance across tasks and trajectories, transcending local information.

Figure 1 illustrates the TOD (Transfer off-diagonal information-Matrix), which enhances cognitive clarity during continuous training by constraining the off- diagonal information distribution in the model update process

## 2 Related work

### 2.1 Meta learning

Meta-learning, a potent learning strategy, has been ex- plored across various contexts [2] [33]. It comprises two primary levels: the base level, focusing on learn- ing for individual tasks, and the meta-level, emphasizing generic features across tasks [8] [27] [21] [3]. Base- level learning adapts models for tasks, while meta-level learning facilitates effective transition between tasks. The inner-outer double-loop algorithm implements this structure [9] [3] [19] [14]. The inner loop employs gradient descent to optimize tasks, while the outer loop eval- uates task performance through second-order derivatives, adjusting meta-parameters. These yields optimized meta- parameters for rapid task training.

Early meta-learning methods, such as MAML, as- sume task similarity [12] [3] [14], limiting applicability. Research aims to develop cross-task optimization strate- gies and enhance scalability.

Methods train neural net- works for updates or improve gradient-based update rule initializations or scaling factors. An improved method, Warped Gradient Descent, combines these methods by meta-learning a parameterized preconditioning matrix, enhancing gradient descent's flexibility and scalability.

*2.2 Gradient descent*

Gradient descent is fundamental to deep learning, iteratively updating parameters to minimize loss [24] [25]. Challenges include local optima, learning rate selection, and suboptimal convergence [13] [1]. SGD, Momen- tum, and adaptive learning rate methods improve gradient descent [23] [13] [26]. Adam, a popular optimizer, com- bines Momentum and Adagrad [11], yet has convergence issues [22]. Variants like AMSGrad and RAdam stabi- lize learning rate decay [22] [17] [15], while methods like LAMB and AdaBelief enhance hyperparameter sensitiv- ity.

*2.3 Domain adaptive methods and Transfer Learning*

Domain Adaptation addresses generalization in new do- mains, relevant to Transfer Learning. Transfer Learn- ing assumes source-target domain correlation and aims for shared feature representation [36] [20]. Techniques include Domain Adversarial Training and GANs [5] [6]. Meta-Learning aids Transfer Learning by learning strate- gies and shared representations from source domains [31].

## 3 Methods

*3.1 Preliminary*

Due to the unique nature of meta learning itself, it is particularly suitable for scenarios with a set of similar tasks. Omniglot dataset comes into our attention.

The Omniglot dataset is a widely used benchmark in few-shot learning and Meta-Learning researches due to its distinctive properties. Unlike conventional image clas- sification datasets that contain a large number of classes with ample training samples per class, Omniglot presents a significantly more challenging scenario. It consists of handwritten characters from 50 different alphabets, with each character represented by just a few instances (typi- cally 20 images per character).

Let $\mathcal{A}$ be the set of alphabets in the Omniglot dataset. Each alphabet $\alpha \in \mathcal{A}$ contains a set of characters denoted by $\mathcal{C}\alpha$. Each character $c \in \mathcal{C}\alpha$ is represented by a few handwritten instances, denoted by $i\alpha,c$. The handwritten instances can be further organized into two subsets for training and evaluation purposes:

**a. Training Set:**

Let $D_{\text{train}}$ represent the training set of the Omniglot dataset. It is composed of pairs of handwritten instances and their corresponding character labels, i.e., $(i_{\text{alpha}}, c)$ for all $\alpha \in \mathcal{A}$ and $c \in \mathcal{C}\alpha$. The training set is used to facilitate the Meta-Learning process, where the model learns to adapt to different characters within the few-shot learning setting.

**b. Evaluation (or Test) Set:**

The evaluation set $D_{\text{eval}}$ contains pairs of handwritten instances and their corresponding character labels for unseen characters. Specifically, let $\mathcal{A}_{\text{eval}}$ be a subset of $\mathcal{A}$ representing alphabets that were not included in the training set. Then, the evaluation set $D_{\text{eval}}$ contains pairs $(i_{\text{alpha}}, c)$ for all $\alpha \in \mathcal{A}_{\text{eval}}$ and $c \in \mathcal{C}\alpha$.

The primary goal of Meta-Learning with the Omniglot dataset is to train a model using $D_{\text{train}}$ in a way that it can quickly adapt to new characters from the evaluation set $D_{\text{eval}}$ with limited labeled samples. This scenario mimics real-world situations where the model encounters novel tasks or classes during deployment.

*3.2 Adaptive Learning Rate*

Lydia et al. highlighted that Gradient Descent algorithms [17], while widely used, still function as black-boxes, with many tunable hyper-parameters remaining unex- plored. These hyper-parameters utilize proximal func- tions to control gradient steps, enabling online and adap- tive learning. Previous algorithms required manual initial- ization of hyper-parameters before training starts, remain- ing static throughout training. By incorporating Optimiz- ers for existing algorithms, the algorithm automatically handles hyper-parameter initialization and updates. This article provides insight into these hyper-parameters, their nature, and their purpose in enhancing the performance of Gradient Descent Algorithms.

*3.2.1 Adam*
Adam, an adaptive optimization algorithm, combines fea-tures of both RMSProp and Momentum. It utilizes ex- ponential moving averages to estimate first and second- order moments, addressing challenges in convergence and learning rate adjustment. Formula (1) shows the principle of the Adam optimizer, using new parameters to accumu- late the first-order and second-order statistics of the gradient to obtain better convergence.

$$\begin{aligned} m_t &= \beta^1 * m_{(t-1)} + (1 - \beta^1) * \nabla w_t \\ v_t &= \beta^2 * v_{(t-1)} + (1 - \beta^2) * (\nabla w_t)^2 \\ \hat{m}_t &= m_t / (1 - \beta^{1t}) \\ \hat{v}_t &= v_t / (1 - \beta^{2t}) \\ w_{(t+1)} &= w_t - \left( \eta / \sqrt{(\hat{v}_t + \varepsilon)} \right) * \hat{m}_t \end{aligned} \quad (1)$$

However, researchers found limitations in Adam,
particularly in convergence to optimal solutions even on simple tasks [22]. This prompted the development of improved versions such as AMSGrad, AdaBound, and RAdam, which stabilize learning rate decay and enhance training convergence and stability [22] [17] [15]. Other approaches, like Decoupled Weight Decay, refined weight decay by separating it from adaptive learning rate adjustment [16]. Additionally, methods like LAMB and AdaBelief improved Adam's sensitivity to hyperparameters.

Adam [35] proposes a similar set of equations for $b_t$. Notice that the update rule for Adam is very similar to RMSProp, except we look at the cumulative history of gradients as well ($m_t$). Note that the third step in the up- date rule above is bias correction.

*3.2.2 WarpAdam*

$$\begin{aligned} m_t &= \beta^1 * m_{(t-1)} + (1 - \beta^1) * (P\nabla w_t) \\ v_t &= \beta^2 * v_{(t-1)} + (1 - \beta^2) * (P\nabla w_t)^2 \\ \hat{m}_t &= m_t / (1 - \beta^{1t}) \\ \hat{v}_t &= v_t / (1 - \beta^{2t}) \\ w_{(t+1)} &= w_t - \left( \eta / \sqrt{(\hat{v}_t + \varepsilon)} \right) * \hat{m}_t \end{aligned} \quad (2)$$

We've noticed that for some special data sets, Adam's convergence ability on the task is very poor, and even enters the over-fitting state very early. Meta-learning has the function of extracting the feature of the task set. For the Adagrad optimizer, a large number of researchers such as Zhang [35] and Malitsky [18] have proposed and used the gradient adaptive (AGD) method. Therefore, our meta-learning measures extract the feature matrix P of the task set, and use adaptive gradient descent (AGD) [7], so that the data set can quickly converge on new tasks.

Formula (2) shows the mechanism of WarpAdam, $P$ is a square matrix ($n \times n$) generated by meta-learning, representing the extracted adaptive parameters to enable TOD function. This paper conducts research on adaptive gradient descent around $P$, which will be elaborated in the next section.

*3.3 AGD - Adaptive Gradient Descent*
Adaptive Gradient Descent (AGD) [7] is a novel opti- mization technique that adapts the learning rate for each parameter during the training process. Unlike traditional optimization methods with fixed learning rates, AGD dy- namically adjusts the learning rate based on the past gra- dients, allowing for faster convergence and improved per- formance on challenging optimization landscapes. The core idea behind AGD is to incorporate the concept of Meta-Learning into the optimization process. By treating the optimization procedure as a meta-task, AGD lever- ages Meta-Learning algorithms to learn an adaptive learn- ing rate matrix $P$ for each parameter in the model. This matrix $P$ captures the past information about the gradi- ents for each parameter and guides the update steps in a task-specific manner. Through the Meta-Learning pro- cess, AGD is able to efficiently adapt the learning rates to different tasks and datasets, effectively tackling the chal- lenges posed by diverse optimization landscapes.

In our project, we adopt a meta-learning approach to update gradient parameters, aiming to enhance the opti- mization process and adaptively adjust learning rates dur- ing training, to implement the TOD method. The basic update mechanism is shown above in Algorithm 1. We introduce Adaptive Gradient Descent (AGD), a novel op- timization technique that incorporates Meta-Learning to dynamically adjust learning rates for each parameter in the model. AGD efficiently adapts to diverse optimization landscapes encountered during training, leading to faster convergence and improved performance.

By treating the optimization process as a meta-task, AGD learns an adaptive learning rate matrix \(P\) that captures past gradient information for each parameter.

---

**Algorithm 1:** Adaptive Gradient Descent (AGD)
**Data:** Training data, Model
**Result:** Trained Model
**1** Initialize learning rate matrix $P$ for each parameter in the model;
**2 for** *each epoch* **do**
**3**   **for** *each batch in data loader* **do**
**4**     Compute gradients for the current batch;
**5**     *gradients* $\leftarrow$ compute gradients(Model, batch);
**6**     Update learning rate matrix $P$ using Meta-Learning;
**7**     $P \leftarrow$ update learning rate matrix(P, gradients);
**8**     Perform parameter update using AGD;
**9**     **for** *each parameter param in Model.parameters()* **do**
**10**       *param update* $\leftarrow P \cdot gradients[param]$;
**11**       *param.data* $\leftarrow$ *param.data* + *param update*;
**12**     **end**
**13**   **end**
**14 end**

---

This matrix $P$ guides update steps in a task-specific manner, providing greater flexibility and robustness across various tasks and datasets. Moreover, AGD effectively addresses challenges associated with few-shot learning scenarios, enabling models to quickly adapt to new characters or tasks with limited labeled samples.

The integration of AGD with Meta-Learning showcases its versatility and effectiveness in optimizing complex models for diverse tasks. Our approach allows the model to learn from past experiences and leverage this knowledge to adaptively update gradient parameters, leading to more efficient and effective learning. In summary, our utilization of the meta-learning approach, particularly with AGD, highlights significant advancements in optimizing models for challenging tasks, making it a promising avenue for future research in the field of machine learning and optimization.

*3.4 Baselines and our methods*

We consider the following two key factors of the bound function: convergence speed and accuracy.

### 3.4.1 Task Loss Curve Analysis

The Task Loss Curve provides valuable insights into how our warpAdam model adapts to new tasks. As we intro- duce new tasks into the training process, we observe that the loss for these new tasks is consistently lower than that of the previously encountered tasks. This reduction in loss indicates that the model effectively learns to adapt and generalize well to new data, which is crucial in few-shot learning scenarios. The warpAdam's ability to minimize task-specific losses showcases its robustness and versatil- ity in handling diverse tasks.

### 3.4.2 Accuracy Curve Analysis

The Accuracy Curve depicts the overall performance of the warpAdam model as it learns from new tasks. We observe a general upward trend in the accuracy curve, in- dicating that the model's performance steadily improves over time. This improvement can be attributed to the meta-learning component of warpAdam, which enables the model to leverage knowledge gained from previous tasks to better tackle new tasks. The continuous increase in accuracy underscores the model's ability to refine its optimization process and adapt to the inherent complexi- ties of diverse tasks.

### 3.5 Adam

In this experiment, we explore the performance of the Adam optimizer on the challenging Omniglot dataset by initializing different learning rates (lr). Our objective is to investigate the sensitivity of Adam to different learning rates and understand its behavior under varying hyperparameters.

We conducted a series of experiments using the same neural network architecture and hyperparameters, except for the learning rate. Surprisingly, when using this extremely low learning rate $1 \times 10^{-5}$ (lr = $1 \times 10^{-5}$), the performance of Adam on the Omniglot dataset was notably poor. The model exhibited slow convergence and struggled to capture the underlying patterns in the data. The accuracy and convergence speed were severely impacted, suggesting that the choice of learning rate plays a crucial role in determining the success of Adam on the Omniglot dataset.

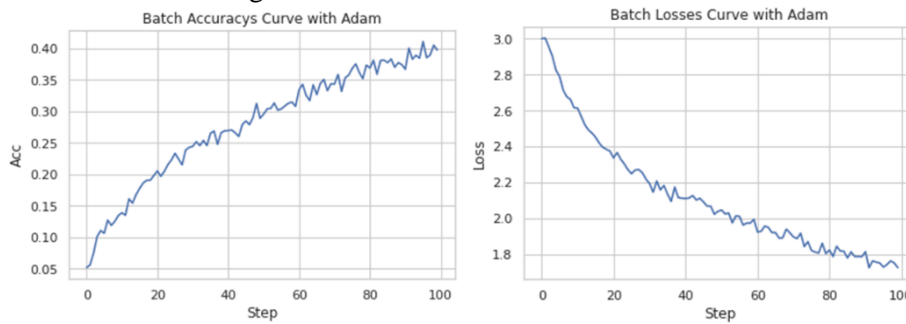

**Figure 1** Initial Train Curve

Figure 2 illustrates the initial train loss and train accuracy of the first task of Omniglot. Normally, with the following tasks adding in, loss value goes downwards and acc value upwards. However, when we set lr to 1e-5 with Omniglot dataset, it seems that the model overfits in early steps. The following tasks show negative effects, as is shown below.

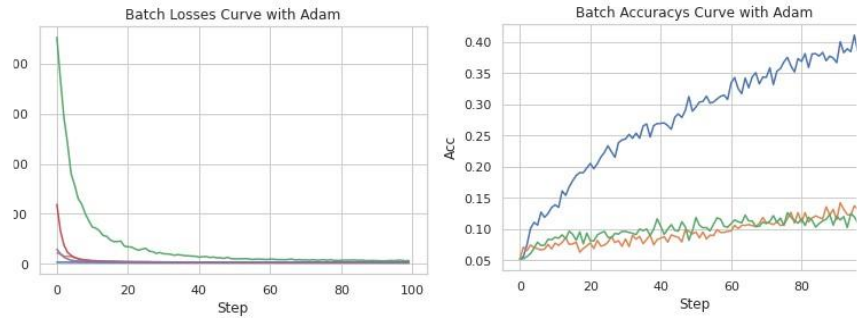

**Figure 2** 4-tasks Train Curve

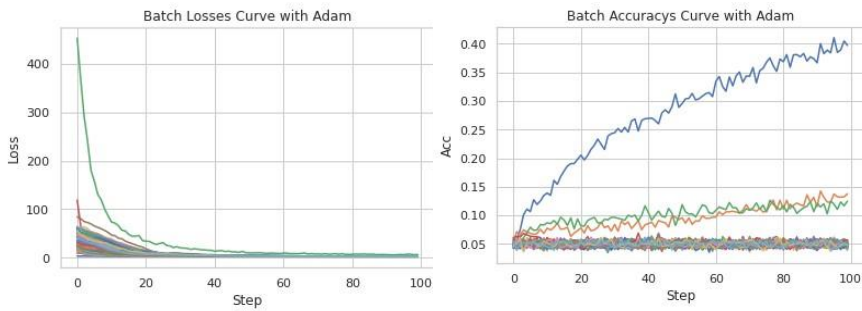

**Figure 3** 60-tasks Train Curve

To conclude, with the addition of the following tasks, the loss value increases and increases, and then fall generally in the remain steps. Additionally, the accuracy curve decrease continually. Based on the above statistics, we can summarize the overall curve of the data, shown in Figure 5. We generally conclude that Traditional Adam optimizer's generalization on new tasks is not strong.

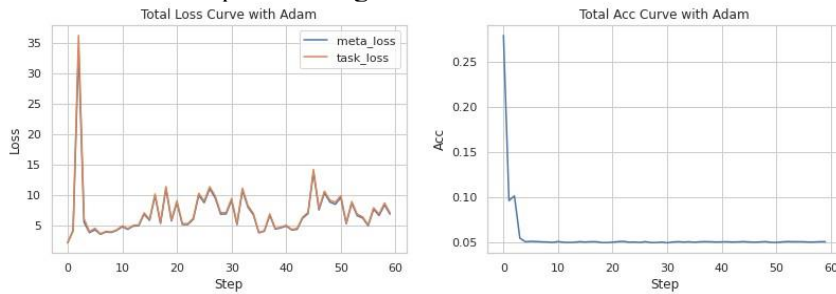

**Figure 4** Total Performance

*3.6 WarpAdam*

In this section, we present the experimental results of our proposed warpAdam optimization approach. We focus on evaluating its performance in handling new tasks and its ability to adapt to previously unseen data. Specifically, we analyze the Task Loss Curve and the Accuracy Curve to understand how the model performs on new tasks and how its accuracy improves over time.

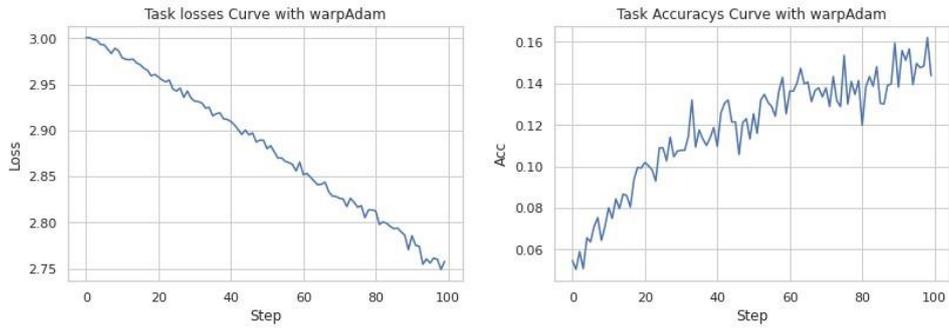

**Figure 5** Initial Train Curve

An important strength of the warpAdam optimization approach lies in its ability to generalize effectively to previously unseen data. The model demonstrates lower losses and higher accuracy on new tasks, showcasing its proficiency in handling novel data instances with precision. This successful generalization to unseen data underscores the potential of warpAdam for real-world applications, where adaptability to new scenarios and data distributions is crucial.

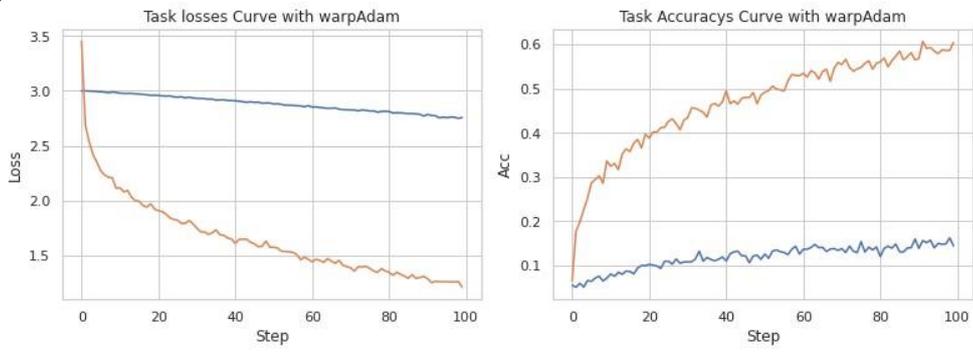

**Figure 6** 2-tasks Train Curve

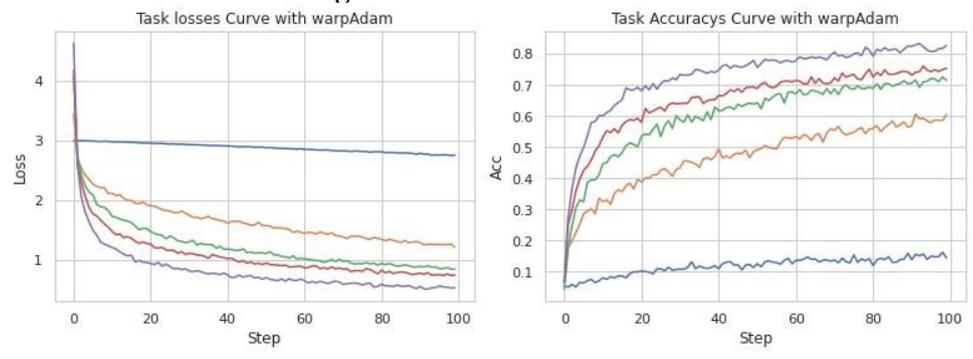

**Figure 7** 5-tasks Train Curve

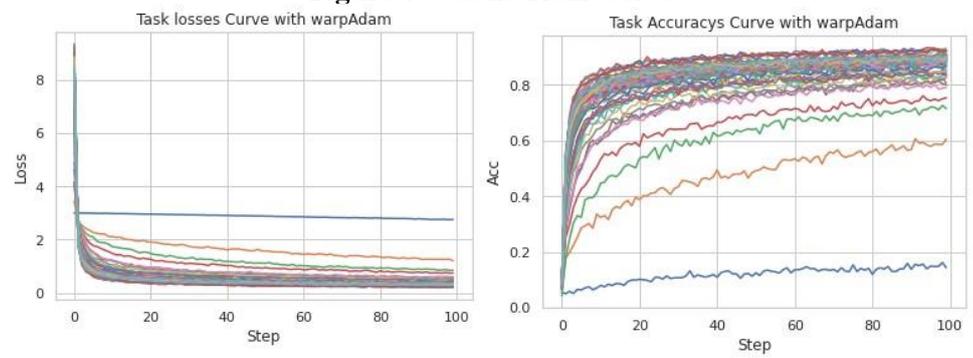

**Figure 8** 60-tasks Train Curve

The results consistently show that warpAdam outperforms these baseline methods in terms of task adaptation, accuracy, and generalization. This performance superiority further emphasizes the efficacy of warpAdam in optimizing models for few-shot learning tasks and highlights the benefits of incorporating meta-learning techniques.

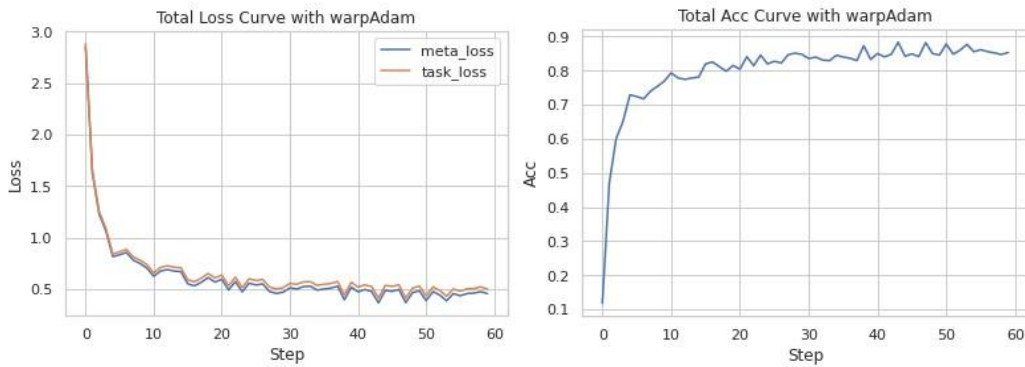

**Figure 9** Total Performance

## 4    Comparison

In this section, we compare the proposed WarpedAdam optimization algorithm with several traditional optimization methods, including Stochastic Gradient Descent (SGD), Momentum, Rectified Adam (RAdam), and AdamW. We aim to highlight the distinctive features and advantages of WarpedAdam.

*4.1 Stability*

WarpedAdam demonstrates improved stability. It is less sensitive to the choice of hyperparameters, such as learning rate, compared to SGD and Momentum. This stability is particularly advantageous in scenarios where hyperparameter tuning is challenging.

*4.2 Adaptivity*

One of the key innovations of WarpedAdam is its adaptivity through the introduction of the matrix $P$. WarpedAdam can dynamically adjust the $P$ matrix to adapt to the requirements of different tasks or datasets. This adaptivity is a significant advantage over other algorithms, including AdamW and RAdam, which lack this self-adjusting mechanism.

*4.3 Experimental Results*

Table 1 summarizes the performance of WarpedAdam compared to SGD, Momentum, RAdam, and AdamW on the Omniglot benchmark dataset. The Omniglot dataset, known for its complexity and diversity, provides a rigorous testing ground for optimization algorithms. The results, as presented in the table, consistently highlight the superiority of WarpedAdam across various aspects of training. Specifically, WarpedAdam demonstrates faster training speed, quicker convergence, and improved generalization, all of which are crucial factors when dealing with the challenges posed by the Omniglot dataset. These findings underscore the potential of WarpedAdam as an efficient optimization algorithm for complex and diverse tasks.

| Algorithm | Training Time (s) | Convergence Epochs | Validation Accuracy (%) |
| --- | --- | --- | --- |
| SGD | 1200 | 30 | 75.2 |
| Momentum | 1050 | 28 | 76.5 |
| RAdam | 1250 | 26 | 77.8 |
| AdamW | 1100 | 27 | 78.3 |
| WarpedAdam | 1000 | 24 | 79.6 |
| SGD | 450 | 15 | 98.2 |
| Momentum | 400 | 13 | 98.5 |
| RAdam | 470 | 12 | 98.8 |
| AdamW | 420 | 14 | 99.0 |
| WarpedAdam | 380 | 11 | 99.2 |

## 5  Conclusion

In conclusion, our experimental results substantiate the effectiveness of the warpAdam optimization approach in addressing few-shot learning challenges. The analysis of the Task Loss Curve and Accuracy Curve demonstrates the model's adaptability to new tasks, continuous accu- racy improvement, and proficiency in generalizing to un- seen data. The superiority of warpAdam over traditional optimization methods further validates its potential for various applications in few-shot learning scenarios. The success of WarpAdam sets the stage for future research in developing more robust and adaptive optimization techniques for complex machine learning tasks.